\documentclass{article}

\usepackage{microtype}
\usepackage{graphicx}
\usepackage{subfigure}
\usepackage{booktabs} % for professional tables
\usepackage{amsmath}
\usepackage{amssymb}
\usepackage{amsthm}
\usepackage{bbm}

\usepackage{hyperref}

\usepackage{graphicx}
\graphicspath{ {./figs/} }

\usepackage{float}

\newtheorem{prop}{Proposition}
\theoremstyle{remark}

\theoremstyle{definition}
\newtheorem*{definition}{Definition}

\newcommand{\R}{\mathbb{R}}
\newcommand{\E}{\mathbb{E}}
\newcommand{\J}{\mathcal{J}}

\newcommand{\normsq}[1]{\left\lVert#1\right\rVert^2}
\newcommand{\bignormsq}[1]{\big\lVert#1\big\rVert^2}

\usepackage[accepted]{icml2019}

\icmltitlerunning{Noise2Self: Blind Denoising by Self-Supervision}

\begin{document}

\twocolumn[
\icmltitle{Noise2Self: Blind Denoising by Self-Supervision}

% unless you've provided the [accepted] option to the icm$L^2$018


\icmlsetsymbol{equal}{*}

\begin{icmlauthorlist}
\icmlauthor{Joshua Batson \textsuperscript{*}}{czb}
\icmlauthor{Loic Royer \textsuperscript{*}}{czb}
\end{icmlauthorlist}

\icmlaffiliation{czb}{Chan-Zuckerberg Biohub}

\icmlcorrespondingauthor{Joshua Batson}{joshua.batson@czbiohub.org}
\icmlcorrespondingauthor{Loic Royer}{loic.royer@czbiohub.org}

\icmlkeywords{Machine Learning, ICML}

\vskip 0.3in
]



\printAffiliationsAndNotice{\icmlEqualContribution} % otherwise use the standard text.

\begin{abstract}
We propose a general framework for denoising high-dimensional measurements which requires no prior on the signal, no estimate of the noise, and no clean training data. The only assumption is that the noise exhibits statistical independence across different dimensions of the measurement, while the true signal exhibits some correlation. For a broad class of functions (``$\mathcal{J}$-invariant''), it is then possible to estimate the performance of a denoiser from noisy data alone. This allows us to calibrate $\mathcal{J}$-invariant versions of any parameterised denoising algorithm, from the single hyperparameter of a median filter to the millions of weights of a deep neural network. We demonstrate this on natural image and microscopy data, where we exploit noise independence between pixels, and on single-cell gene expression data, where we exploit independence between detections of individual molecules. This framework generalizes recent work on training neural nets from noisy images and on cross-validation for matrix factorization.
\end{abstract}

\section{Introduction}
\label{introduction}

We would often like to reconstruct a signal from high-dimensional measurements
that are corrupted, under-sampled, or otherwise noisy. Devices like high-resolution cameras, electron microscopes, 
and DNA sequencers are capable of producing measurements in the thousands
to millions of feature dimensions. But when these devices are pushed to their limits,
taking videos with ultra-fast frame rates at very low-illumination, 
probing individual molecules with electron microscopes, or sequencing tens of thousands of cells simultaneously, 
each individual feature can
become quite noisy. Nevertheless, the objects being studied are often very structured
and the values of different features are highly correlated. Speaking loosely,
if the ``latent dimension" of the space of objects under study
 is much lower than the dimension of the measurement,
 it may be possible to implicitly learn that structure,
 denoise the measurements, and recover the signal without any prior knowledge of the signal or the noise.

Traditional denoising methods each exploit a property of the noise, such as 
Gaussianity, or structure in the signal,  
such as spatiotemporal smoothness, self-similarity, or having low-rank. The performance of these
methods is limited by the accuracy of their assumptions. For example,
if the data are genuinely not low rank, then a low rank model will
fit it poorly. This requires prior knowledge
of the signal structure, which limits application to new domains and modalities.
These methods also require calibration, as hyperparameters
such as the degree of smoothness, the scale of self-similarity, or the rank
of a matrix have dramatic impacts on performance. 

In contrast, a data-driven prior, such as pairs $(x_i, y_i)$ of noisy and clean measurements of
the same target, can be used to set up a supervised learning problem.
A neural net trained to predict $y_i$ from $x_i$ may be used to denoise
new noisy measurements \cite{weigert_content-aware_2018}. 
As long as the new data are drawn from the same distribution,
one can expect performance similar to that observed during training.
Lehtinen et al. demonstrated that clean targets are unnecessary \yrcite{lehtinen_noise2noise:_2018}. 
A neural net trained on pairs $(x_i, x^\prime_i)$ of \emph{independent} noisy measurements
of the same target will, under certain distributional assumptions, learn
to predict the clean signal. These supervised approaches extend to image denoising the success
 of convolutional neural nets, which currently give state-of-the-art
performance for a vast range of image-to-image tasks.
Both of these methods require an experimental setup
in which each target may be measured multiple times, which can be difficult in practice.

\begin{figure*}
  \centering
  %\begin{mdframed}[linecolor=white!30,backgroundcolor=black!5]
    \includegraphics[scale=1]{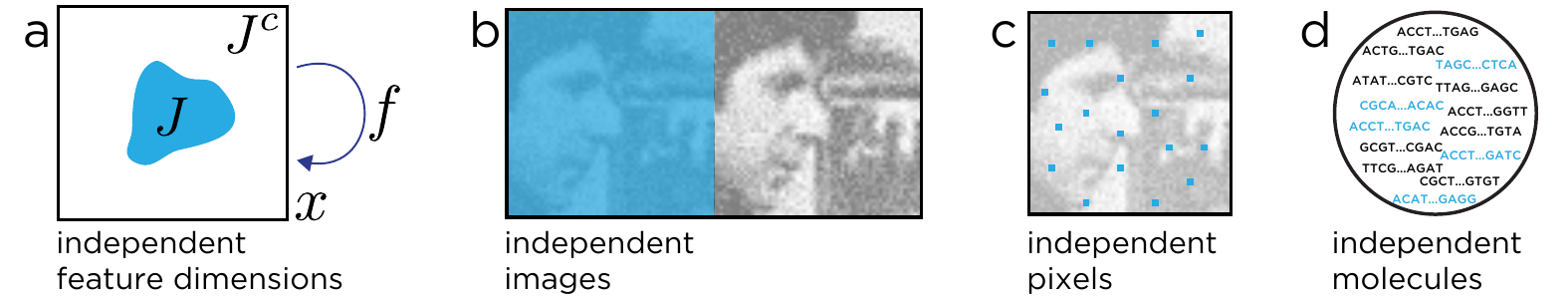}
  %\end{mdframed}
  \caption{(a) The box represents the dimensions of the measurement $x$. $J$ is a subset of the dimensions, and $f$ is a $J$-invariant function: it has the property that the value of $f(x)$ restricted to dimensions in $J$, $f(x)_J$, does not depend on the value of $x$ restricted to $J$, $x_J$. This enables self-supervision when the noise in the data is conditionally independent between sets of dimensions. Here are 3 examples of dimension partitioning: (b) two independent image acquisitions, (c) independent pixels of a single image, (d) independently detected RNA molecules from a single cell.}
  \label{fig:method}
\end{figure*}

In this paper, we propose a framework for blind denoising based on \emph{self-supervision}.
We use groups of features whose noise is independent conditional on the
true signal to predict one another. This allows us to learn 
denoising functions from single noisy measurements of each object,
with performance close to that of supervised methods.
The same approach can also be used to calibrate traditional
image denoising methods such as median filters and non-local means, and, using 
a different independence structure, denoise highly under-sampled single-cell gene expression data.

We model the signal $y$ and its noisy measurement $x$ as a pair of random 
variables in $\R^m$. If $J \subset \{1, \dots , m\}$ is a subset of 
the dimensions, we write $x_J$ for $x$ restricted to $J$.

\begin{definition}
Let $\J$ be a partition of the dimensions $\{1, \dots, m\}$ and let $J\in \J$.
A function $f:\R^m \rightarrow \R^m$ is \emph{$J$-invariant} if $f(x)_J$ does not 
depend on the value of $x_J$. It is \emph{$\J$-invariant} if it is \emph{$J$-invariant} 
for each $J \in \J$.
\end{definition}

We propose minimizing the self-supervised loss
\begin{equation}\label{eqn:n2s_loss}
    \mathcal{L}(f) = \E \normsq{f(x) - x},
\end{equation}
over $\J$-invariant functions $f$. Since $f$ has to use 
information from outside of each subset of dimensions 
$J$ to predict the values inside of $J$,
it cannot merely be the identity.

\begin{prop}\label{prop:ss-equality} Suppose $x$ is an unbiased estimator of $y$, i.e. $\E [x|y] = y$, and the noise in each subset
$J \in \J$ is independent from the noise in its complement $J^c$,
conditional on $y$. Let $f$ be $\J$-invariant. Then
\begin{equation}\label{eqn:loss_decomposition}
\E \normsq{f(x) - x} = \E \normsq{f(x) - y} + \E \normsq{x - y}.
\end{equation}
\end{prop}
That is, the self-supervised loss is the sum of the ordinary supervised loss and the variance of the noise. \emph{By minimizing the 
self-supervised loss over a class of $\J$-invariant functions, one may find the optimal denoiser for a given dataset.}

For example, if the signal is an image with independent, mean-zero
noise in each pixel, we may choose $\mathcal{J} = \{ \{1\},\dots,\{m\}\}$ to be the singletons
 of each coordinate. Then ``donut" median filters, with a hole in the center,
form a class of $\J$-invariant functions, and by comparing the value of the self-supervised loss at different filter radii, we are able to select the optimal radius for denoising the image at hand (See \S 3).

The donut median filter has just one parameter and therefore limited 
ability to adapt to the data. At the other extreme, we may
search over \emph{all} $\J$-invariant functions for the global optimum:
\begin{prop}\label{prop:n2s} The $\J$-invariant function
$f^*_\J$ minimizing \eqref{eqn:n2s_loss} satisfies
$$f^*_\J(x)_J = \mathbb{E} [y_J \vert x_{J^c}]$$
for each subset $J \in \J$.
\end{prop}
That is, the optimal $\J$-invariant predictor for the dimensions of 
$y$ in some $J \in \J$ is their 
expected value conditional on observing the dimensions of $x$ outside 
of $J$.

In \S 4, we use analytical examples to illustrate how the optimal 
$\J$-invariant denoising function approaches the optimal general 
denoising function as the amount of correlation between features 
in the data increases.

In practice, we may attempt to approximate the optimal denoiser by
searching over a very large class of functions, such as 
deep neural networks with millions of parameters. In \S 5, we show that a deep convolutional network,
modified to become $\J$-invariant 
using a masking procedure, can achieve state-of-the-art blind denoising
performance on three diverse datasets.

Sample code is available on GitHub\footnote{\url{https://github.com/czbiohub/noise2self}} 
and deferred proofs are contained in the Supplement.

\section{Related Work}

Each approach to blind denoising relies on assumptions about the structure of the signal 
and/or the noise. We review the major categories of assumption below, and the traditional 
and modern methods that utilize them. Most of the methods below are described in terms 
of application to image denoising, which has the richest literature, but some have 
natural extensions to other spatiotemporal signals and
to generic measurements of vectors.

\textbf{Smoothness:} Natural images and other spatiotemporal signals are often
assumed to vary smoothly \cite{buades_review_2005}. Local averaging, using a Gaussian,
median, or some other filter, is a simple way to smooth out a noisy input. The degree 
of smoothing to use, e.g., the width of a filter, is a hyperparameter often tuned by 
visual inspection.

\textbf{Self-Similarity:} Natural images are often self-similar, in that each
patch in an image is similar to many other patches from the same image. 
The classic non-local means algorithm replaces the center pixel of each patch with a 
weighted average of central pixels from similar patches \cite{buades_non-local_2005}. The more robust 
BM3D algorithm makes stacks of similar
patches, and performs thresholding in frequency space \cite{dabov_image_2007}. 
The hyperparameters of these methods have a 
large effect on performance \cite{lebrun_analysis_2012}, and on a new dataset with an unknown noise distribution it is difficult to evaluate their effects in a principled way.

Convolutional neural nets can produce images with another form of
self-similarity, as linear combinations of the same small filters
are used to produce each output. The ``deep image prior" of \cite{ulyanov_deep_2017}
exploits this by training a generative CNN to produce a single output image 
and stopping training before the net fits the noise.

\textbf{Generative:} Given a differentiable, generative model of the data, e.g. a neural net
$G$ trained using a generative adversarial loss, data can be denoised through projection 
onto the range of the net \cite{tripathi_correction_2018}.

\textbf{Gaussianity:} Recent work \cite{zhussip_training_2018, metzler_unsupervised_2018} uses a 
loss based on Stein's unbiased risk estimator to train denoising neural nets in the special case that noise is i.i.d. Gaussian.

\textbf{Sparsity}: Natural images are often close to sparse in e.g. a wavelet or DCT basis \cite{chang_adaptive_2000}. 
Compression algorithms such as JPEG exploit this feature
by thresholding small transform coefficients \cite{pennebaker_jpeg_1992}. This is also a denoising
strategy, but artifacts familiar from poor compression (like the
ringing around sharp edges) may occur. Hyperparameters include the choice of basis and the degree of thresholding. Other methods
learn an overcomplete dictionary from the data and seek sparsity in that basis \cite{elad_image_2006, papyan_convolutional_2017}.

\textbf{Compressibility}: A generic approach to denoising
is to lossily compress and then decompress the data. The accuracy of this approach
depends on the applicability of the compression scheme used to the
signal at hand and its robustness to the form of noise. It also depends
on choosing the degree of compression correctly: too much will lose
important features of the signal, too little will preserve all of the noise. 
For the sparsity methods, this ``knob'' is the degree of sparsity, while for 
low-rank matrix factorizations, it is the rank of the matrix.

Autoencoder architectures for neural nets provide a general framework
for learnable compression. Each sample is mapped to a low-dimensional
representation---the value of the neural net at the bottleneck layer---
then back to the original space \cite{gallinari_memoires_1987, vincent_stacked_2010}.
An autoencoder trained on noisy data may produce cleaner data
as its output. The degree of compression is determined by the width of
the bottleneck layer.

UNet architectures, in which skip connections are added to a typical
autoencoder architecture, can capture high-level spatially coarse representations 
and also reproduce fine detail; they can, in particular,
learn the identity function \cite{ronneberger_u-net:_2015}. Trained directly on noisy data, they
will do no denoising. Trained with clean targets, they can learn
very accurate denoising functions \cite{weigert_content-aware_2018}.

\textbf{Statistical Independence:} Lehtinen et al. observed that a
UNet trained to predict one noisy measurement of a signal from an 
independent noisy measurement of the same signal will 
in fact learn to predict the true signal \cite{lehtinen_noise2noise:_2018}. We may reformulate the Noise2Noise procedure in terms of $\J$-invariant functions: if 
$x_1 = y + n_1$ and $x_2 = y + n_2$ are the two measurements, we consider the 
composite measurement $x = (x_1, x_2)$ of a composite signal $(y, y)$ in $\R^{2m}$
and set $\mathcal{J} = \{J_1, J_2\} = \{\{1,\dots,m\},\{m+1,\dots,2m\}\}$.
Then $f^*_\J (x)_{J_2} = \E [y | x_1]$.

\begin{figure*}[!ht]
  \centering
  %\begin{mdframed}[linecolor=white!30,backgroundcolor=black!5]
    \includegraphics[scale=1]{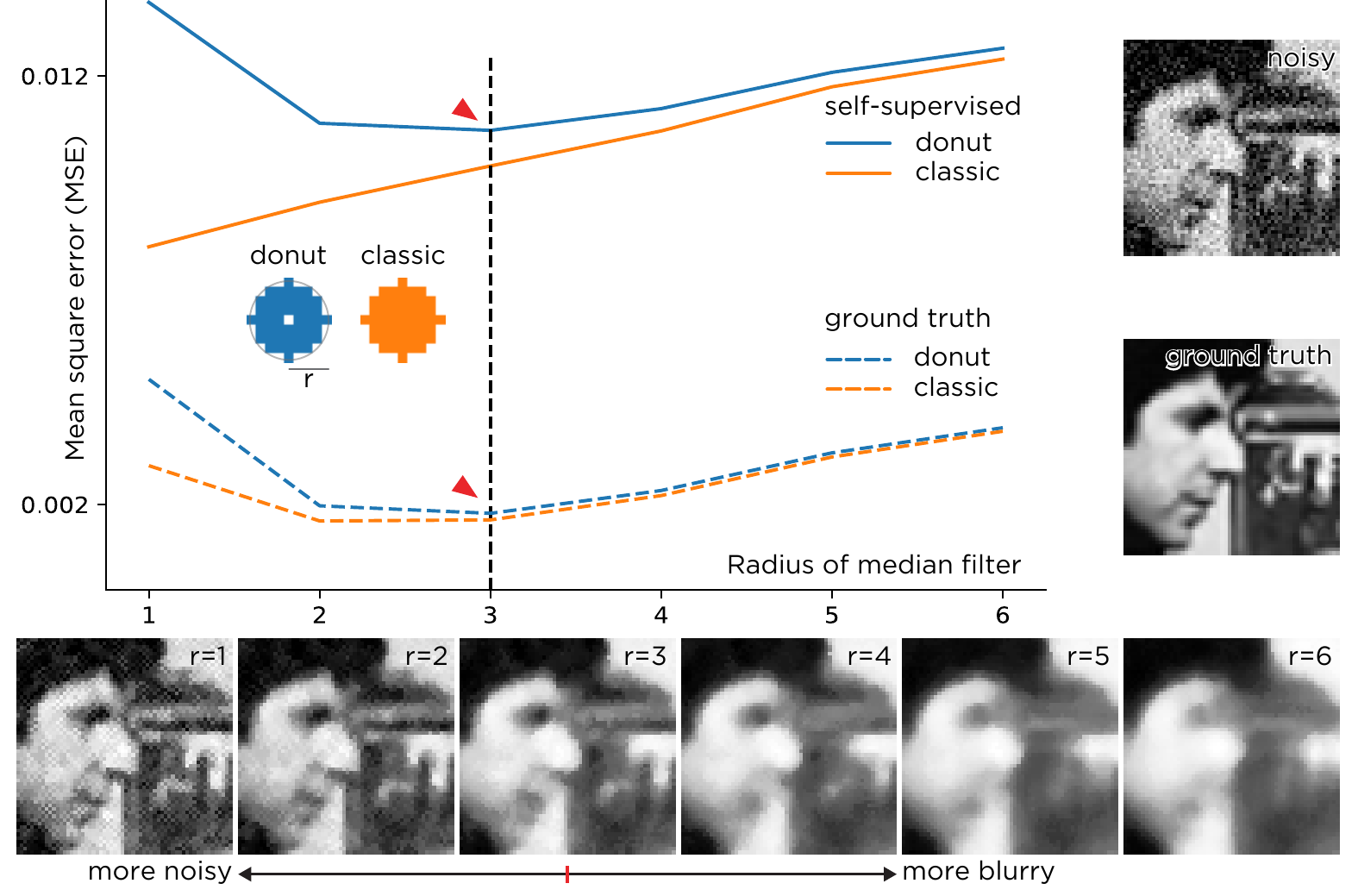}
  %\end{mdframed}
  \caption{Calibrating a median filter without ground truth. Different median filters may be obtained by varying the filter's radius. Which is optimal for a given image? The optimal parameter for $\J$-invariant functions such as the donut median can be read off (red arrows) from the self-supervised loss.    
  }
  \label{fig:calibration}
\end{figure*}

An extension to video, in which one frame is used to compute the pullback under 
optical flow of another, was explored in \cite{ehret_model-blind_2018}.

In concurrent work, Krull et al. train a UNet to predict a collection of held-out pixels of an image from 
a version of that image with those pixels replaced \yrcite{krull_noise2void_2018}. A key difference
between their approach and our neural net examples in \S 5 is 
in that their replacement strategy is not quite $\J$-invariant. (With 
some probability a given pixel is replaced by itself.) While
their method lacks a theoretical guarantee against fitting the noise, it 
performs well in practice, on natural and
microscopy images with synthetic and real noise.

Finally, we note that the ``fully emphasized denoising autoencoders" in \cite{vincent_stacked_2010}
used the MSE between an autoencoder evaluated on masked 
input data and the true value of the masked pixels, but with 
the goal of learning robust representations, not denoising.

\section{Calibrating Traditional Models}\label{section:calibration}
Many denoising models have a hyperparameter controlling the degree of the denoising---the size of a filter, the threshold for sparsity, the number of principal components. If ground truth data were available, the optimal parameter $\theta$ for a family of denoisers $f_\theta$ could be chosen by minimizing $\normsq{f_\theta(x) - y}$. Without ground truth, we may nevertheless compute the self-supervised loss $\normsq{f_\theta(x) - x}$. For general $f_\theta$, it is unrelated to the ground truth loss, but if $f_\theta$ is $\J$-invariant, then it is equal to the ground truth loss plus the noise variance (Eqn. \ref{eqn:loss_decomposition}), and will have the same minimizer.

In Figure~\ref{fig:calibration}, we compare both losses for the median filter $g_r$, which replaces each pixel with the median over a disk of radius $r$ surrounding it, and the ``donut'' median filter $f_r$, which replaces each pixel with the median over the same disk \emph{excluding the center}, on an image with i.i.d. Gaussian noise. For $\J = \{\{1\}, \dots, \{m\}\}$ the partition into single pixels, the donut median is $\J$-invariant. For the donut median, the minimum of the self-supervised loss $\normsq{f_r(x) - x}$ (solid blue) sits directly above the minimum of the ground truth loss $\normsq{f_r(x) - y}$ (dashed blue), and selects the optimal radius $r = 3$. The vertical displacement is equal to the variance of the noise. In contrast, the self-supervised loss $\normsq{g_r(x) - x}$ (solid orange) is strictly increasing and tells us nothing about the ground truth loss $\normsq{g_r(x) - y}$ (dashed orange). Note that the median and donut median are genuinely different functions with slightly different performance, but while the former can only be tuned by inspecting the output images, the latter can be tuned using a principled loss.

More generally, let $g_\theta$ be any classical denoiser, and let $\J$ be any partition of the pixels such that neighboring pixels are in different subsets. Let $s(x)$ be the 
function replacing each pixel with the average of its neighbors. Then the function $f_\theta$ defined by
\begin{equation}\label{eqn:mask}
f_\theta(x)_J := g_\theta(\mathbf{1}_{J}\cdot s(x) + \mathbf{1}_{J^c}\cdot x)_J,
\end{equation}
for each $J \in \J$, is a $\J$-invariant version of $g_\theta$. Indeed, since the pixels of $x$ in $J$ are replaced before applying $g_\theta$, the output cannot depend on $x_J$.

In Supp. Figure~1, we show the corresponding loss curves for $\J$-invariant versions of a wavelet filter, where we tune the threshold $\sigma$, and NL-means, where we tune a cut-off distance $h$ \cite{buades_non-local_2005, chang_adaptive_2000, van_der_walt_scikit-image:_2014}. The partition $\mathcal{J}$ used is a 4x4 grid. Note that in all these examples, the function $f_\theta$ is genuinely different than $g_\theta$, and, because the simple interpolation procedure may itself be helpful, it sometimes performs better.

In Table \ref{table:comparing-trad-models}, we compare all three $\J$-invariant denoisers on a single image. As expected, the denoiser with the best self-supervised loss also has the best performance as measured by Peak Signal to Noise Ratio (PSNR).

\begin{table}[t]
\caption{Comparison of optimally tuned $\J$-invariant versions of classical denoising models. Performance is better than original method at default parameter values, and can be further improved (+) by adding an optimal amount of the noisy input to the $\J$-invariant output (\S \ref{section:doingbetter}).}
\label{table:comparing-trad-models}
\vskip 0.15in
\begin{center}
\begin{small}
\begin{sc}
\begin{tabular}{lcccc}
\toprule
Method & Loss &  & PSNR & \\
 & J-invt & J-invt & J-invt+ & Default\\

\midrule
 Median   & 0.0107 & 27.5 & 28.2 & 27.1 \\ 
 Wavelet        & 0.0113  & 26.0 & 26.9 & 24.6 \\ 
 NL-Means       & \textbf{0.0098} & \textbf{30.4} &\textbf{30.8} & \textbf{28.9}  \\
\bottomrule
\end{tabular}
\end{sc}
\end{small}
\end{center}
\vskip -0.1in
\end{table}

\subsection{Single-Cell} In single-cell transcriptomic experiments, thousands of individual cells are isolated, lysed, and their mRNA are extracted, barcoded, and sequenced. Each mRNA molecule is mapped to a gene, and that $\sim$20,000-dimensional vector of counts is an approximation to the gene expression of that cell. In modern, highly parallel experiments, only a few thousand of the hundreds of thousands of mRNA molecules present in a cell are successfully captured and sequenced \cite{milo_bionumbers_2010}. Thus the expression vectors are very undersampled, and genes expressed at low levels will appear as zeros. This makes simple relationships among genes, such as co-expression or transitions during development, difficult to see.

If we think of the measurement as a set of molecules captured from a given cell, then we may partition the molecules at random into two sets $J_1$ and $J_2$. Summing (and normalizing) 
the gene counts in each set produces expression vectors $x_{J_1}$ and $x_{J_2}$ which are independent conditional on the true mRNA content $y$. We may now attempt to denoise $x$ by training 
a model to predict $x_{J_2}$ from $x_{J_1}$ and vice versa.

\begin{figure}[h!]
  \centering
  %\begin{mdframed}[linecolor=white!30,backgroundcolor=black!5]
    \includegraphics[scale=1]{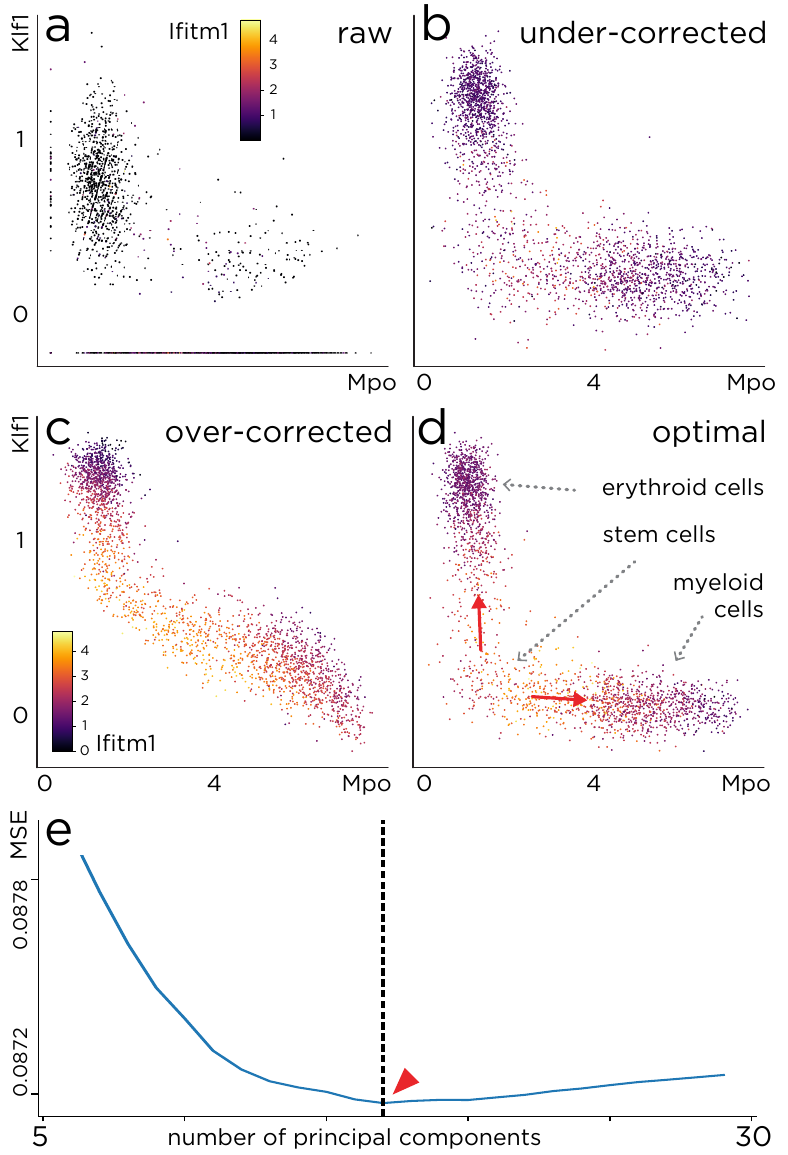}
  %\end{mdframed}
  \caption{Self-supervised loss calibrates a linear denoiser for single cell data. 
  (a) Raw expression of three genes: a myeloid cell marker (Mpo), an erythroid cell marker (Klf1), and a stem cell marker (Ifitm1). Each point corresponds to a cell. (e) Self-supervised loss for principal component regression. In (d) we show the  the denoised data for the optimal number of principal components (17, red arrow). In (c) we show the result of using too few components and in (b) that of using too many. X-axes show square-root normalised counts.
  }
  \label{fig:singlecell}
\end{figure}

We demonstrate this on a dataset of 2730 bone marrow cells from Paul et al. using principal component regression \cite{paul_transcriptional_2015},
where we use the self-supervised loss to find an optimal number of principal components. The data contain a population of stem cells which differentiate either into erythroid or myeloid lineages. The expression of genes preferentially expressed in each of these cell types is shown in Figure~\ref{fig:singlecell} for both the (normalized) noisy data and data denoised with too many, too few, and an optimal number of principal components. In the raw data, it is difficult to discern any population structure. When the data is under-corrected, the stem cell marker \textit{Ifitm1} is still not visible. When it is over-corrected, the stem population appears to express substantial amounts of \textit{Klf1} and \textit{Mpo}. In the optimally corrected version, \textit{Ifitm1} expression coincides with low expression of the other markers, identifying the stem population, and its transition to the two more mature states is easy to see.

\subsection{PCA} Cross-validation for choosing the rank of a PCA requires some care, since 
adding more principal components will always produce a better fit, even on held-out samples \cite{bro_cross-validation_2008}. Owen and Perry recommend splitting the feature dimensions into two sets $J_1$ and $J_2$ as well as splitting the samples into train and validation sets \cite{owen_bi-cross-validation_2009}. For a given $k$, they fit a rank $k$ principal component regression $f_k: X_{\text{train},J_1} \mapsto X_{\text{train},J_2}$ and evaluate its predictions on the validation set, computing $\normsq{f_k(X_{\text{valid},J_1}) - X_{\text{valid},J_2}}$. They repeat this, permuting train and validation sets and $J_1$ and $J_2$. Simulations show that if $X$ is actually a sum of a low-rank matrix plus Gaussian noise, then the $k$ minimizing the total validation loss is often the optimal choice \cite{owen_bi-cross-validation_2009, owen_bi-cross-validation_2016}. This calculation corresponds to using the self-supervised loss to train and cross-validate a $\{J_1 ,J_2 \}$-invariant principal component regression.

\section{Theory}\label{section:theory}

In an ideal situation for signal reconstruction, we have
a prior $p(y)$ for the signal and a probabilistic model of the
noisy measurement process $p(x|y)$. After observing some
measurement $x$, the posterior distribution for $y$ is
given by Bayes' rule:
$$p(y|x) = \frac{p(x|y)p(y)}{\int p(x|y)p(y)dy}.$$
In practice, one seeks some function $f(x)$ approximating 
a relevant statistic of $y|x$, such as its mean or median. The
mean is provided by the function minimizing the loss:
$$\E_x \normsq{f(x) - y}$$
(The $L^1$ norm would produce the median) \cite{murphy_machine_2012}.

Fix a partition $\J$ of the dimensions $\{1, \dots, n \}$ of $x$ and suppose that for
each $J \in \J$, we have
$$p(x|y) = p(x_J|y)p(x_{J^c}|y),$$
i.e., $x_J$ and $x_{J^c}$ are independent conditional on $y$. We consider the loss
\begin{align*}
\E_x \normsq{f(x) - x} = \E_{x,y} &\normsq{f(x) - y} + \normsq{x - y} \\ 
                           & - 2 \langle f(x) - y, x - y \rangle.
\end{align*}
If $f$ is $\J$-invariant, then for each $j$ the random variables $f(x)_j|y$ and $x_j|y$ are independent. The third term reduces to $\sum_j \E_y ( \E_{x|y}[f(x)_j - y_j])(\E_{x|y} [x_j - y_j])$, which 
vanishes when $\E [x | y] = y$. This proves Prop. \ref{prop:ss-equality}.

Any $\J$-invariant function can be written as a collection
of ordinary functions $f_J: \R^{\vert J^c \vert} \rightarrow \R^{\vert J \vert}$, where 
we separate the output dimensions of $f$ based on which input dimensions they depend on.
Then 
$$\mathcal{L}(f) = \sum_{J \in \J} \E \normsq{f_J(x_{J^c}) - x_J}.$$
This is minimized at
\begin{align*}
    f^*_J(x_{J^c})  &= \E [x_J \vert x_{J^c}] = \E [y_J|x_{J^c}].
\end{align*}
We bundle these functions into $f^*_\J$, proving Prop. \ref{prop:n2s}.

\begin{figure}[th]
  \centering
  %\begin{mdframed}[linecolor=white!30,backgroundcolor=black!5]
    \includegraphics[scale=1]{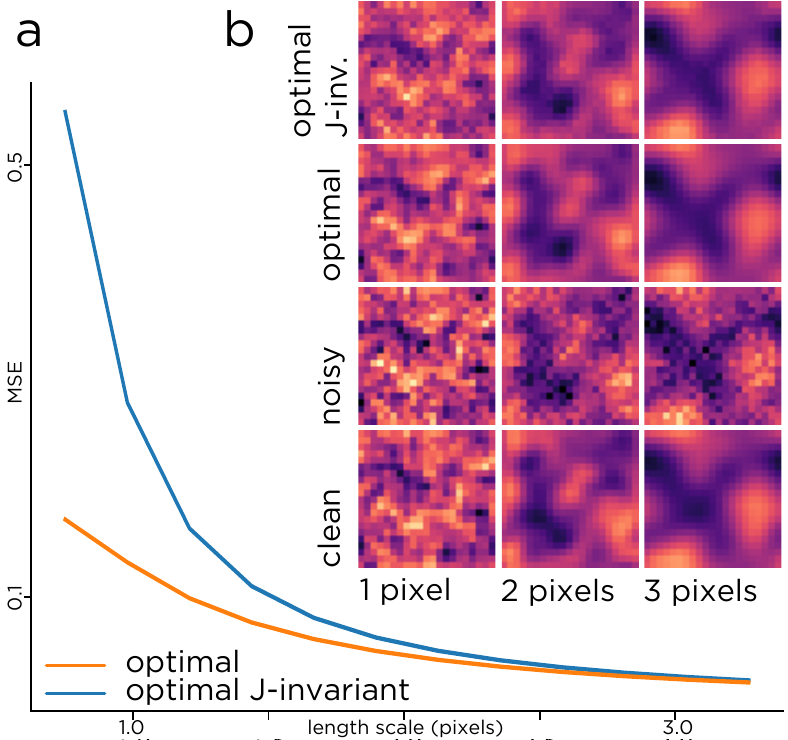}
  %\end{mdframed}
  \caption{The optimal $\J$-invariant predictor converges to the optimal predictor. Example images for Gaussian processes of different length scales. The gap in image quality between the two predictors tends to zero as the length scale increases.}
  \label{fig:theory1}
\end{figure}

\begin{figure*}[ht]
  \centering
  %\begin{mdframed}[linecolor=white!30,backgroundcolor=black!5]
    \includegraphics[scale=1]{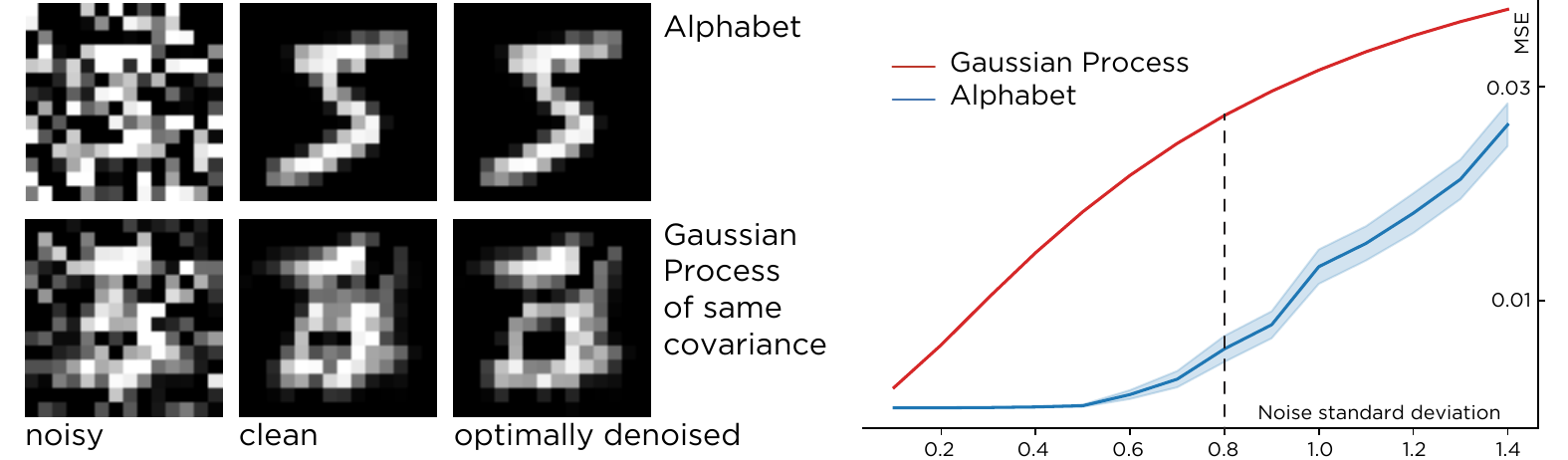}
  %\end{mdframed}
  \caption{For any dataset, the error of the optimal predictor (blue) is lower than that for a Gaussian Process (red) with the same covariance matrix. We show this for a dataset of noisy digits: the quality of the denoising is visibly better for the Alphabet than the Gaussian Process (samples at $\sigma=0.8$).   
  }
  \label{fig:theory2}
\end{figure*}

\subsection{How good is the optimum?}

How much information do we lose by giving up $x_J$ when trying to predict $y_J$?
Roughly speaking, the more the features in $J$ are correlated with those outside of it,
the closer $f^*_\J(x)$ will be to $\E [y|x]$ and the better both will estimate $y$.

Figure~\ref{fig:theory1} illustrates this phenomenon for the example of Gaussian Processes, 
a computationally tractable model of signals with correlated features. We consider a 
process on a $33 \times 33$ toroidal grid.  The value of $y$ at each node 
is standard normal and the correlation between the values at $p$ and $q$ depends on the distance between them: 
$K_{p, q} = \exp(-\normsq{p - q}/2\ell^2)$, where $\ell$ is the length scale. 
The noisy measurement $x = y + n$, where $n$ is white Gaussian noise with standard deviation $0.5$.

While 
$$\E \bignormsq{y - f^*_\J(x)} \geq \E \bignormsq{y - \E [y|x]}$$
for all $\ell$, the gap decreases quickly as the length scale increases.

The Gaussian process is more than a convenient example; it actually represents 
a worst case for the recovery error as a function of correlation.
\begin{prop}
Let $x, y$ be random variables and let $x^G$ 
and $y^G$ be Gaussian random variables with the same covariance matrix. Let
$f^*_{\J}$ and $f^{*,G}_{\J}$ be the corresponding optimal $\J$-invariant 
predictors. Then
$$\E \bignormsq{y - f^*_{\J}(x)} \leq \E \bignormsq{y - f^{*,G}_{\J}(x)}.$$
\end{prop}
\begin{proof}
See Supplement.
\end{proof}
Gaussian processes represent a kind of local texture with no 
higher structure, and the functions $f^{*,G}_\J$ turn out to be linear \cite{murphy_machine_2012}.

At the other extreme is data drawn from finite collection of templates, like symbols in an alphabet. If the alphabet consists of $\{a_1,\dots,a_r\} \in \R^m$ and the noise is i.i.d. mean-zero Gaussian with variance $\sigma^2$, then the optimal $J$-invariant prediction independent is a weighted sum of the letters from the alphabet. The weights
$w_i = \exp(-\normsq{(a_i - x)\cdot \mathbf{1}_{J^c}}/2\sigma^2)$
 are proportional to the posterior probabilities of each letter.
When the noise is low, the output concentrates on a copy of the closest letter; when the noise is high, the output averages many letters.

In Figure~\ref{fig:theory2}, we demonstrate this phenomenon for an alphabet consisting of 30 16x16 handwritten 
digits drawn from MNIST \cite{lecun_gradient-based_1998}. Note that almost exact 
recovery is possible at much higher levels of noise than the Gaussian process with covariance matrix given by the empirical covariance matrix of the alphabet. Any real-world dataset will exhibit more structure than a Gaussian process, so nonlinear functions can generate significantly better predictions.

\subsection{Doing better}\label{section:doingbetter}

If $f$ is $\J$-invariant, then by definition $f(x)_j$ contains no information from $x_j$, and the right linear combination $\lambda f(x)_j + (1 - \lambda) x_j$ will produce an estimate of $y_j$ with lower variance than either. The optimal value of $\lambda$ is given by the variance of the noise divided by the value of the
self-supervised loss. The performance gain depends on the quality of $f$: for example, if $f$ improves the PSNR by 10 dB, then mixing in the optimal amount of $x$ will yield another 0.4 dB. (See Table~\ref{table:comparing-trad-models} for an example and Supplement for proofs.)

\section{Deep Learning Denoisers}\label{section:experiments}

\begin{figure*}[!ht]
  \centering
  %\begin{mdframed}[linecolor=white!30,backgroundcolor=black!5]
    \includegraphics[scale=1]{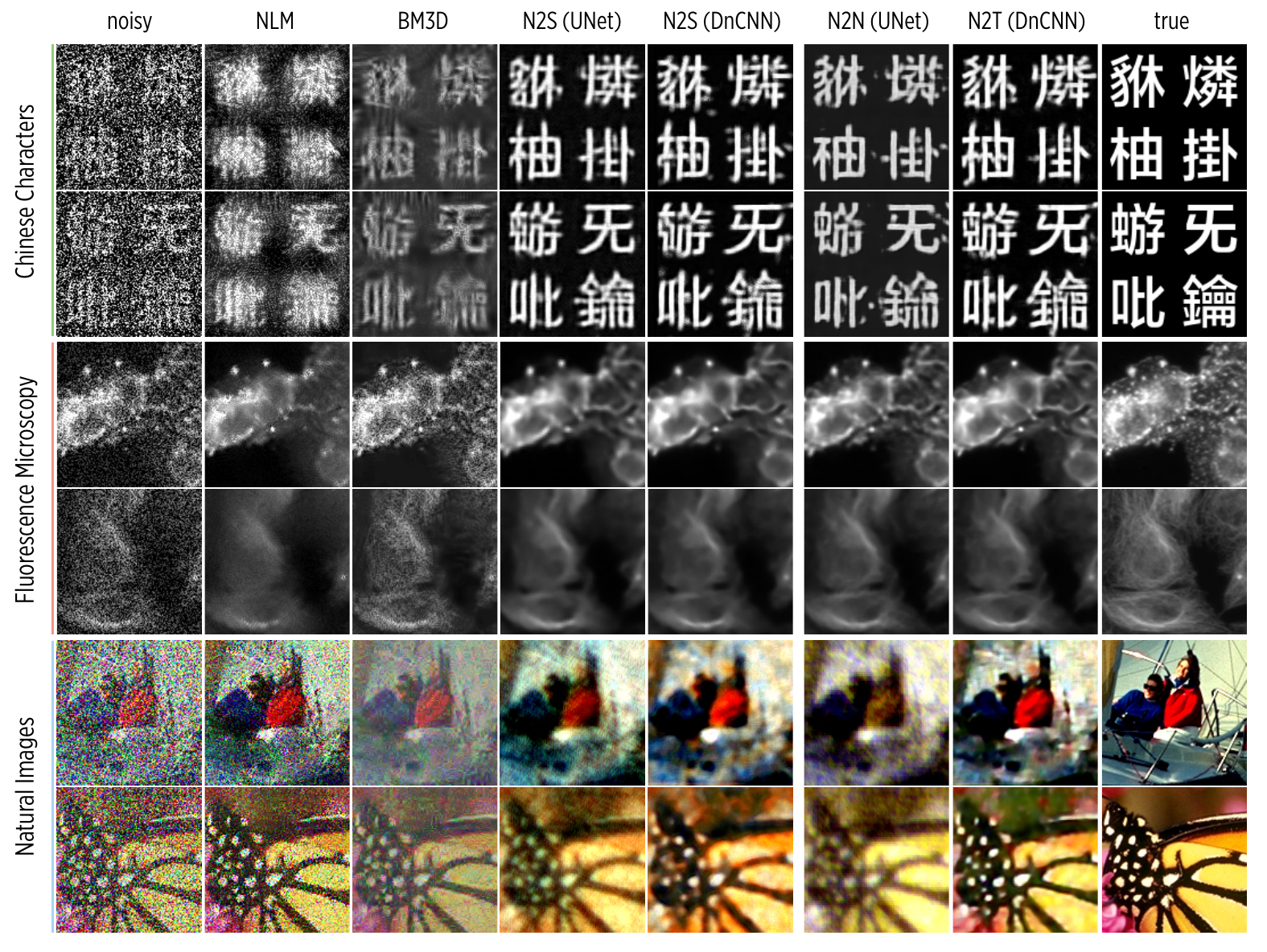}
  %\end{mdframed}
  \caption{Performance of classic, supervised, and self-supervised denoising methods on natural images, Chinese characters, and fluorescence microscopy images. Blind denoisers are NLM, BM3D, and neural nets (UNet and DnCNN) trained with self-supervision (N2S). We compare to neural nets supervised with a second noisy image (N2N) and with the ground truth (N2T).}
  \label{fig:experiments}
\end{figure*}

The self-supervised loss can be used to train a deep convolutional neural net with just one noisy sample of each image in a dataset. We show this on three datasets from different domains (see Figure~\ref{fig:experiments}) with strong and varied heteroscedastic synthetic noise applied independently to each pixel. For the datasets \textbf{H\`anz\`i} and \textbf{ImageNet} we use a mixture of Poisson, Gaussian, and Bernoulli noise. For the \textbf{CellNet} microscopy dataset we simulate realistic sCMOS camera noise.  We use a random partition of 25 subsets for $\J$, and we make the neural net $\J$-invariant as in Eq.~\ref{eqn:mask}, except we replace the masked pixels with random values instead of local averages. We train two neural net architectures, a UNet and a purely convolutional net, DnCNN \cite{zhang_beyond_2017}. To accelerate training, we only compute the net outputs and loss for one partition $J \in \J$ per minibatch.

As shown in Table~\ref{table:experiment-results}, both neural nets trained with self-supervision (Noise2Self) achieve superior performance to the classic unsupervised denoisers NLM and BM3D (at default parameter values), and comparable performance to the same neural net architectures trained with clean targets (Noise2Truth) and with independently noisy targets (Noise2Noise).

The result of training is a neural net $g_\theta$, which, when converted into a $\J$-invariant function $f_\theta$, has low self-supervised loss.
We found that applying $g_\theta$ directly to the noisy input gave slightly better ($0.5$ dB) performance than using $f_\theta$. The images in Figure~\ref{fig:experiments} use $g_\theta$.

Remarkably, it is also possible to train a deep CNN to denoise a single noisy image. The DnCNN architecture, with ~560,000 parameters, trained with self-supervision on the noisy \texttt{camera} image from \S3, with ~260,000 pixels, achieves a PSNR of 31.2.

\section{Discussion}

We have demonstrated a general framework for denoising high-dimensional measurements whose noise exhibits some conditional independence structure. We have shown how to use a self-supervised loss to calibrate or train any $\J$-invariant class of denoising functions.

There remain many open questions about the optimal choice of partition $\J$ for a given problem. The structure of $\J$ must reflect the patterns of dependence in the signal and independence in the noise. The relative sizes of each subset $J\in \J$ and its complement creates a bias-variance tradeoff in the loss, exchanging information used to make a prediction for information about the quality of that prediction.

\begin{table}[]
\caption{Performance of different denoising methods by Peak Signal to Noise Ratio (PSNR) on held-out test data. Error bars for CNNs from training five models.}
\label{table:experiment-results}
\vskip 0.15in
\begin{center}
\begin{small}
\begin{sc}
\begin{tabular}{lccc}
\toprule
Method & H\`anz\`i & ImageNet & CellNet \\ [0.5ex] 
\midrule
 Raw         & 6.5  & 9.4  & 15.1 \\ 
 NLM         & 8.4 & 15.7 & 29.0 \\ 
 BM3D        & 11.8 & 17.8 & 31.4 \\
 UNet \quad\;(n2s) & 13.8 $\pm$ 0.3 & 18.6 & 32.8 $\pm$ 0.2\\
 DnCNN (n2s) & 13.4 $\pm$ 0.3 & 18.7 & 33.7 $\pm$ 0.2\\
\midrule
 UNet  \quad\;(n2n) & 13.3 $\pm$ 0.5 & 17.8 & 34.4 $\pm$ 0.1\\
 DnCNN (n2n) & 13.6 $\pm$ 0.2 & 18.8 & 34.4 $\pm$ 0.1\\
\midrule
 UNet  \quad\;(n2t) & 13.1 $\pm$ 0.7 & 21.1 & 34.5 $\pm$ 0.1\\
 DnCNN (n2t) & 13.9 $\pm$ 0.6 & 22.0 & 34.4 $\pm$ 0.4\\
\bottomrule
\end{tabular}
\end{sc}
\end{small}
\end{center}
\vskip -0.1in
\end{table}

For example, the measurements of single-cell gene expression could be partitioned by molecule, gene, or even pathway, reflecting different assumptions about the kind of stochasticity occurring in transcription.

We hope this framework will find application to other domains, such as sensor networks in agriculture or geology, time series of whole brain neuronal activity, or telescope observations of distant celestial bodies.

\section*{Acknowledgements}

Thank you to James Webber, Jeremy Freeman, David Dynerman, Nicholas Sofroniew, 
Jaakko Lehtinen, Jenny Folkesson, Anitha Krishnan, and Vedran Hadziosmanovic for valuable conversations. 
Thank you to Jack Kamm for discussions on Gaussian Processes and shrinkage estimators.
Thank you to Martin Weigert for his help running BM3D. Thank you to the referees for suggesting valuable clarifications. 
Thank you to the Chan Zuckerberg Biohub for financial support.

\nocite{van_dijk_recovering_2018}
\nocite{ljosa_annotated_2012}
\nocite{paszke_automatic_2017}

\bibliography{references_cleaned}
\bibliographystyle{icml2019}

\end{document}

\twocolumn[

\icmltitle{Supplement to Noise2Self: Blind Denoising by Self-Supervision}
]

\section{Notation}

For a variables $x \in \R^m$ and $J \subset \{ 1, \dots, m \}$, we write $x_J$ for the restriction of $x$ to 
the coordinates in $J$ and $x_{J^c}$ for the restriction of $x$ to the coordinates in $J^c$. If $f: \R^m \rightarrow \R^m$
is a function, we write $f(x)_J$ for the restriction of $f(x)$ to the coordinates in $J$.

A partition $\J$ of a set $X$ is a set of disjoint subsets of $X$ whose union is all of $X$.

When $J = \{j\}$ is a singleton, we write $x_{-j}$ for $x_{J^c}$, the restriction of $x$ to the coordinates not equal to $j$.

\section{Gaussian Processes}

Let $x$ and $y$ be random variables. Then the estimator of $y$ from $x$ minimizing the 
expected mean-square error (MSE) is 
$x \mapsto \E [y|x]$. The expected MSE of that estimator is simply the variance of $y|x$:

\[ \E_x \normsq{y - \E [y|x]} = \E_x \var(y|x). \]

If $x$ and $y$ are jointly multivariate normal, then the right-hand-side depends only 
on the covariance matrix $\Sigma.$ If 

\[
\Sigma=
\left(
\begin{array}{cc}
\Sigma_{xx} & \Sigma_{yx} \\
\Sigma_{xy} & \Sigma_{yy}
\end{array}
\right),
\]

then then right-hand-side is in fact a constant independent of $x$:

$$\var(y|x) = \Sigma_{yy} - \Sigma_{yx}\Sigma_{xx}^{-1}\Sigma_{xy}.$$

(See Chapter 4 of \cite{murphy_machine_2012}.)

\begin{lemma}\label{lemma:psd}
Let $\Sigma$ be a symmetric, positive semi-definite matrix with block structure 

\[
\Sigma=
\left(
\begin{array}{cc}
\Sigma_{11} & \Sigma_{12} \\
\Sigma_{21} & \Sigma_{22}
\end{array}
\right)
.\]
Then \[\Sigma_{11} \succeq \Sigma_{12} \Sigma_{22}^{-1} \Sigma_{21}.\]
\end{lemma}
\begin{proof}
Since $\Sigma$ is PSD, we may factorize it as a product $X^TX$ for some matrix $X$. 
(For example, take the spectral decomposition $\Sigma = V^T \Lambda V$, with $\Lambda$ the 
diagonal matrix of eigenvalues, all of which are nonnegative since $\Sigma$ is PSD and 
$V$ the matrix of eigenvectors. Set $X$ = $\Lambda^{1/2} V$.)

Write 

\[
X=
\left(
\begin{array}{cc}
X_1 & X_2
\end{array}
\right)
,\]

so that $\Sigma_{ij} = X_i^T X_j$. If $\pi_{X_2}$ is the projection 
operator onto the column-span of $X_2$, then

\begin{align*}
    I       &\succeq \pi_{X_2}  = X_2 (X_2^T X_2)^{-1} X_2^T.
\end{align*}

Multiplying on the left and right by $X_1^T$ and $X_1$ yields

\begin{align*}
   X_1^T X_1   &\succeq X_1^T X_2 (X_2^T X_2)^{-1} X_2^T X_1 \\
\Sigma_{11} &\succeq \Sigma_{12} \Sigma_{22}^{-1} \Sigma_{21},
\end{align*}
where the second line follows by grouping terms in the first.
\end{proof}

\begin{lemma}\label{lemma:guassian-cov}
Let $x, y$ be random variables and let $x^G$ 
and $y^G$ be Gaussian random variables with the same covariance matrix.
Then 

$$\E_x \normsq{y -\E [y | x]} \leq \E_{x^G} \normsq{y^G -\E [y^G | x^G]}.$$
\end{lemma}

\begin{proof}
These are in fact the expected variances of the conditional variables:

$$\E \normsq{y - \E y | x} = \E_x \E_{y|x} \normsq{y - \E y | x} = \E_x \var[y|x].$$

Using the formula above for the Gaussian process MSE, we now need to show that

$$\E_x \var[y|x] \leq \Sigma_{yy} - \Sigma_{yx}\Sigma_{xx}^{-1}\Sigma_{xy}.$$

By the law of total variance, 

\[\var (y) = \var_x(\E [y|x]) + \E_x \var(y|x). \]

So it suffices to show that 

$$\var_x(\E [y|x]) \geq \Sigma_{yx}\Sigma_{xx}^{-1}\Sigma_{xy}.$$

Without loss of generality, we set $\E x = \E y = 0$. We compute the covariance 
of $x$ with $\E [y|x]$. We have 

\begin{align*}
    \operatorname{Cov}(x, \E [y|x]) &= \E_x \left [ x \cdot \E \left [ y|x\right ]\right] \\
                                  &= \E_x \left [ \E \left [ xy|x\right ]\right] \\
                                  &= \E_x \left [ \E \left [ xy|x\right ]\right] \\
                                  &= \E [xy] \\
                                  &= \operatorname{Cov}(x,y)
\end{align*}

The statement follows from an application of Lemma~\ref{lemma:psd} to 
the covariance matrix of $x$ and $\E[y|x]$.
\end{proof}

\begin{prop}
Let $x, y$ be random variables and let $x^G$ 
and $y^G$ be Gaussian random variables with the same covariance matrix. Let
$f^*_{\J}$ and $f^{*,G}_{\J}$ be the corresponding optimal $\J$-invariant 
predictors. Then

$$\E \normsq{y - f^*_{\J}(x)} \leq \E \normsq{y - f^{*,G}_{\J}(x)}.$$
\end{prop}

\begin{proof}
We first reduce the statement to unconstrained optimization, noting that 

$$f^*_{\J}(x)_j = \E y_j | x_{J^c}.$$

The statement follows from Lemma~\ref{lemma:guassian-cov} applied to $y_j, x_{J^c}$.
\end{proof}

\section{Masking}

In this section, we discuss approaches to modifying the input to a neural 
net or other function $f$ to create a $\J$-invariant function.

The basic idea is to choose some interpolation function $s(x)$ and then 
define $g$ by

$$g(x)_J := f(\mathbf{1}_{J}\cdot s(x) + \mathbf{1}_{J^c}\cdot x)_J,$$

where $\mathbf{1}_J$ is the indicator function of the set $J$.

In Section 3 of the paper, on calibration, $s$ is given by a local average, not
containing the center. Explicitly, it is convolution with the kernel 

\[
\left(
\begin{array}{ccc}
0 & 0.25 & 0 \\
0.25 & 0 & 0.25 \\
0 & 0.25 & 0 
\end{array}
\right).
\]

We also considered setting each entry of $s(x)$ to a random variable 
uniform on $[0,1]$. This produces a random $\J$-invariant function,
ie, a distribution $g(x)$ whose marginal $g(x)_J$ does not depend on
$x_J$.

\subsection{Uniform Pixel Selection}

In Krull et. al., the authors propose masking procedures that estimate a 
local distribution $q(x)$ in the neighborhood of a pixel and then 
replace that pixel with a sample from the distribution. Because the 
value at that pixel is used to estimate the distribution, information about
it leaks through and the resulting random functions are not genuinely $J$-invariant.

For example, they propose a method called Uniform Pixel Selection (UPS) 
to train a neural net to predict $x_j$
from $\UPS_j(x)$, where $\UPS_j$ is the random function replacing 
the $j^{th}$ entry of $x$ with the value of at a pixel $k$ chosen uniformly 
at random from the $r \times r$ neighborhood centered at $j$ \cite{krull_noise2void_2018}.

Write $\iota_{jk}(x)$ is the vector $x$ with the value $x_j$ replaced by $x_k$.

The function $f^*$ minimizing the self-supervised loss 

$$\E_x \normsq{f(\UPS_j(x))_j - x_j}$$

satisfies

\begin{align*}
f^*(x)_j &= \E_x[x_j|\UPS_j(x)] \\ 
             &= \E_x\;\E_k [x_j|\iota_{jk}(x)] \\ 
             &= \frac{1}{r^2} \sum_k \E[x_j|\iota_{jk}(x)] \\
             &= \frac{1}{r^2} \E[x_j|\iota_{jj}(x)] + \frac{1}{r^2} \sum_{k \neq j} \E[x_j|\iota_{jk}(x)] \\
             &= \frac{1}{r^2} x_j + \frac{1}{r^2} \sum_{k \neq j} \E[x_j|x_{-j}] \\
             &= \frac{1}{r^2} x_j + \Big (1 - \frac{1}{r^2} \Big) f^*_\J(x)_j,
\end{align*}

where $f^*_\J(x)_j = \E [x_j | x_{-j}]$ is the optimum of the self-supervised loss among $\J$-invariant functions.

This means that training using UPS masking can, given sufficient data and a sufficiently expressive network,
produce a linear combination of the noisy input and the Noise2Self optimum. 
The smaller the region used for selecting the pixel, the larger the contribution of the noise will be.
In practice, however, a convolutional neural net may not be able to learn to recognize when it was handed an 
interesting pixel $x_j$ and when it had been replaced (say by comparing the  value at a pixel in $\UPS_j(x)$ to each of its neighbors).

One attractive feature of UPS is that it keeps the same per-pixel data distribution as the input. If, for example,
the input is binary, then local averaging and random uniform replacements will both be substantial 
deviations. This may regularize the behavior of the network,
making it more sensible to pass in an entire copy of $x$ to the trained network later, rather than iteratively masking it.

We suggest a simple modification: exclude the value of $x_j$ when estimating the local distribution.
For example, replace it with a random neighbor.

\subsection{Linear Combinations}

In this section we note that if $f$ is $\J$-invariant, then $f(x)_j$ and $x_j$ give two uncorrelated estimators of $y_j$ for any coordinate $j$.
Here we investigate the effect of taking a linear combination of them.

Given two uncorrelated and unbiased estimators $u$ and $v$ of some quantity $y$, we may form a linear combination:
    
$$w_\lambda = \lambda u + (1 - \lambda)v.$$

The variance of this estimator is

$$\lambda^2 U  + (1 - \lambda)^2 V,$$

where $U$ and $V$ are the variances of $u$ and $v$ respectively. This expression is minimized at

$$\lambda = V/(U + V).$$

The variance of the mixed estimator $w_\lambda$ is $UV/(U+V) = V\frac{1}{1+V/U}$. When the variance of $v$ is much lower than that of $u$, we just get $V$ out, but when they are the same the variance is exactly halved. Note that this is monotonic in $V$, so if estimators $v_1, \dots, v_n$ are being compared, their rank will not change after the original signal is mixed in. In terms of PSNR, the new value is

\begin{align*}
\PSNR(w_\lambda, y) &= 10*\log_{10}\left (\frac{1+V/U}{V} \right) \\
                    &= \PSNR(V) + 10*\log_{10}(1 + V/U) \\
                    &\approx \PSNR(V) + \frac{10}{\log_{10}(e)} \left (\frac{V}{U} - \frac{1}{2}\frac{V^2}{U^2} \right ) \\
                    &\approx \PSNR(V) + 4.3 \cdot \frac{V}{U}
\end{align*}

If we fix $y$, then $x_j$ and $\mathbb{E} [y_j|x_{-j}]$ are both independent estimators of $y_j$, so the above reasoning applies. 
Note that the loss itself is the variance of $x_j|x_{-j}$, whose two components are the variance of $x_j|y_j$ and the variance of $y_j|x_{-j}$.

The optimal value of $\lambda$, then, is given by the variance of the noise divided by the value of the self-supervised loss. 
For example the function $f$ reduces the noise by a factor of 10 (ie, the variance of $y_j | x_{-j}$ is a tenth of the variance of $x_j | y_j$), then $\lambda^* = 1/11$ and the linear combination has a PSNR $0.43$ higher than that of $f$ alone.

\section{Calibrating Traditional Denoising Methods}

The image denoising methods were all demonstrated on the full \texttt{camera} image 
included in the \texttt{scikit-image} library for python \cite{van_der_walt_scikit-image:_2014}.
An inset from that image was displayed in the figures.

We also used the \texttt{scikit-image} implementations of the median filter, wavelet denoiser, and NL-means.
The noise standard deviation was $0.1$ on a $[0,1]$ scale.

In addition to the calibration plots for the median filter in the text, we show the same for the wavelet and NL-means denoisers in Supp. Figure 1.

\begin{figure*}[!ht]
  \centering
  %\begin{mdframed}[linecolor=white!30,backgroundcolor=black!5]
    \includegraphics[scale=1]{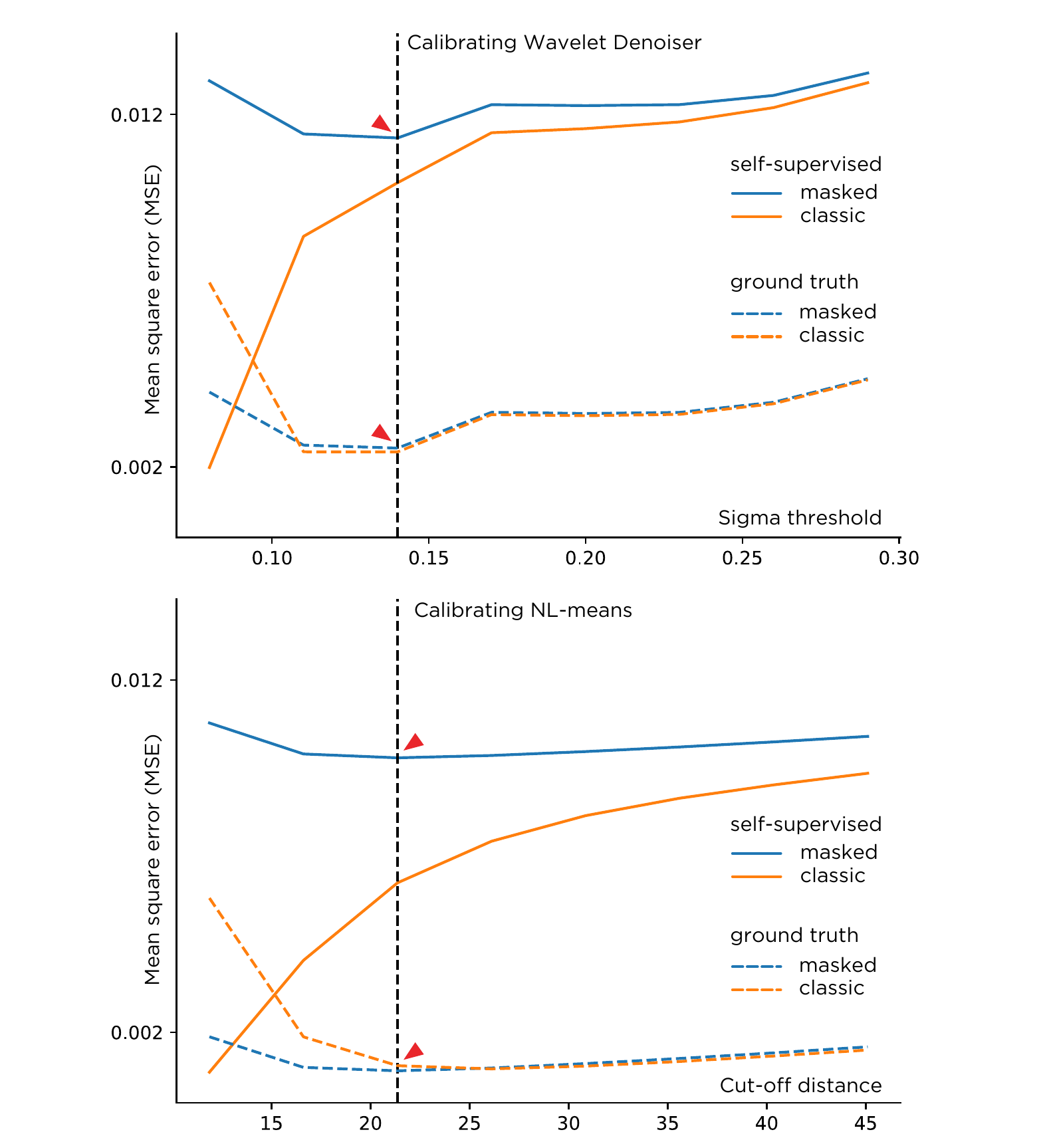}
  %\end{mdframed}
  \caption{Calibrating a wavelet filter and Non-local means without ground truth. The optimal parameter for $\J$-invariant (masked) 
  versions can be read off (red arrows) from the self-supervised loss.
  }
  \label{fig:singlecell}
\end{figure*}

\section{Neural Net Examples}

\subsection{Datasets: H\`anz\`i, CellNet, ImageNet}

\textbf{H\`anz\`i} We constructed a dataset of 13029 Chinese characters (h\`anz\`i) rendered as white on black 64x64 images (image intensity within $[0,1]$), and applied to each one substantial Gaussian ($\mu=0, \sigma=0.7$) and Bernoulli (half pixels blacked out) noise. Each Chinese character appears 6 times in the whole dataset of 78174 images. We then split this dataset in a training and test set (90\% versus 10\%). 

\textbf{CellNet} We constructed a dataset of 34630 image tiles (128x128) obtained by random partitioning of a large collection of single channel fluorescence microscopy images of cultured cells. These images were downloaded from the \emph{Broad Bioimage Benchmark Collection} \cite{ljosa_annotated_2012}. Before cropping, we first gently denoise the images using the non-local means algorithm. We do so in order to remove a very \emph{low} and nearly imperceptible amount of noise already present in these images -- indeed, the images have an excellent signal-to-noise ratio to start from. Next, we use a rich noise model to simulate typical noise on sCMOS scientific cameras. This noise model consists of: (i) spatially variant gain noise per pixel, (ii) Poisson noise, (iii) Cauchy distributed additive noise. We choose parameters so as to obtain a very aggressive noise regime.  

\textbf{ImageNet} In order to generate a large collection of natural image tiles, we downloaded the \emph{ImageNet LSVRC 2013 Validation Set} consisting of 20121 RGB images -- typically photographs. From these images we generated 60000 cropped images of dimension 128x128 with each RGB value within $[0,255]$. These images were mistreated by the strong combination of Poisson ($\lambda=30$), Gaussian ($\sigma=80$), and Bernoulli noise ($p=0.2$). In the case of Bernoulli noise, each pixel channel (R, G, or B) has probability $p$ of being dark or hot, i.e. set to the value $0$ or $255$.  

\subsection{Architecture}

We use a UNet architecture modelled after \cite{ronneberger_u-net:_2015}. The network has an hourglass shape with skip connections between layers of the same scale. 
Each convolutional block consists of two convolutional layers with 3x3 filters followed by an InstanceNorm. The number of channels is [32, 64, 128, 256]. Downsampling uses strided convolutions and upsampling uses transposed convolutions. The network is implemented in PyTorch \cite{paszke_automatic_2017} and the code is also included in the supplement.

\subsection{Training}

We convert a neural net $f_\theta$ into a random $\J$-invariant function:

\begin{equation}\label{eqn:J-invt-f}
\sum_{J \in \J} \mathbf{1}_J \cdot f_\theta(\mathbf{1}_{J^c} \cdot x_J + \mathbf{1}_J \cdot u)
\end{equation}

where $u$ is a vector of random numbers distributed uniformly on $[0, 1]$. To speed up training, we only compute the coordinates for one $J$ 
per pass, and that $J$ is chosen randomly for each batch with density $1/25$. The loss is restricted to those coordinates.

We train with a batch size of 64 for $\textbf{H\`anz\`i}$ and $\textbf{CellNet}$ and a batch size of 32 for $\textbf{ImageNet}$.

We train for 50 epochs for $\textbf{CellNet}$, 30 epochs for $\textbf{H\`anz\`i}$ and 1 epoch for $\textbf{ImageNet}$.

\subsection{Inference}

We considered two approaches for inference. In the first, we consider a partition $\J$ containing 25 sets and apply Equation (\ref{eqn:J-invt-f})
to produce a genuinely $\J$-invariant function. This requires $|\J|$ applications of the network.

In the second, we just apply the trained network to the full noisy data. This will include the information from $x_j$ in the prediction
$f_\theta(x)_j$. While the information in this pixel was entirely redundant during training, some regularization induced by the 
convolutional structure of the net and the training procedure may have caused it to learn a function which uses that information 
in a sensible way. Indeed, on our three datasets, the direct application was about 0.5 dB better than the $\J$-independent version.

\subsection{Evaluation} 

We evaluated each reconstruction method using the Peak Signal-to-Noise Ratio (PSNR). For two images with range $[0, 1]$, this is 
a log-transformation of the mean-squared error:

$$\operatorname{PSNR}(x, y) = 10*\log_{10}(1/\normsq{x - y}).$$

Because of clipping, the noise on the image datasets is not conditionally mean-zero. (Any noise on a pixel with intensity $1$, for example, must 
be negative.) This induces a bias: $\E [x | y]$ is shrunk slightly towards the mean intensity. For methods trained with clean targets, like Noise2Truth and DnCNN, this effect doesn't matter; the network can learn to produce the correct value. The outputs of the blind methods like Noise2Noise, Noise2Self, NL-means, and BM3D, will exhibit this shrinkage. To make up for this difference, we rescale the outputs of all methods to match the mean and variance of the ground truth.

We compute the PSNR for fully reconstructed images on hold-out test sets which were not part of the training or validation procedure.

\section{Single-Cell Gene Expression}

The lossy capture and sequencing process producing single-cell gene expression can be expressed as a Poisson distribution \footnote{
While the polymerase chain reaction (PCR) used to amplify the molecules for sequencing would introduce random multiplicative distortions,
many modern datasets introduce unique molecular indentifiers (UMIs), barcodes attached to each molecule before amplification 
which can be used to deduplicate reads from the same original molecule.}. A given cell 
has a density $\lambda = (\lambda_1, \dots, \lambda_m)$ over genes $i \in \{1, \dots m\}$, with $\sum_i \lambda_i = 1$.
If we sample $N$ molecules, we get a multinomial distribution which can be approximated as $x_i \sim \operatorname{Poisson}(N\lambda_i)$.

While one would like to model molecular counts directly, the large dynamic range of gene expression (about 5 orders of magnitude) makes
linear models difficult to fit directly. Instead, one typically introduces a normalized variable $z$, for example 

$$z_i = \rho (N_0 * x_i/N),$$

where $N = \sum_i x_i$ is the total number of molecules in a given cell, $N_0$ is a normalizing constant, and $\rho$ is some nonlinearity. Common values for $\rho$ include $x \mapsto \sqrt{x}$ and $x \mapsto \log(1 + x)$.

Our analysis of the Paul et al. dataset \cite{paul_transcriptional_2015} follows one from the tutorial for a diffusion-based denoiser called \href{http://nbviewer.jupyter.org/github/KrishnaswamyLab/magic/blob/master/python/tutorial_notebooks/bonemarrow_tutorial.ipynb}{MAGIC},
and we use the \texttt{scprep} package to perform normalization \cite{van_dijk_recovering_2018}. In the language above, $N_0$ is the median of the total molecule count per cell 
and $\rho$ is square root.

Because we work on the normalized variable $z$, the optimal denoiser would predict

$$ \E [z_i | \lambda] \approx \E_{x_i \sim \operatorname{Poisson}{N \lambda_i}} \sqrt{x_i}\sqrt{N_0/N}.$$

This function of $\lambda_i$ is positive, monotonic and maps $0$ to $0$, so it is directionally informative.
Since expectations do not commute with nonlinear functions, inverting it would not produce an unbiased estimate of $\lambda_i$.
Nevertheless, it provides a quantitative estimate of gene expression which is well-adapted to the large dynamic range.

\bibliography{references_cleaned}
\bibliographystyle{icml2019}